# Quantum Computation via Sparse Distributed Representation

Gerard J. Rinkus

**ABSTRACT**

Quantum superposition states that any physical system simultaneously exists in *all* of its possible states, the number of which is exponential in the number of entities composing the system. The strength of presence of each possible state in the superposition—i.e., the probability with which it would be observed if measured—is represented by its probability amplitude coefficient. The assumption that these coefficients must be represented physically disjointly from each other, i.e., *localistically*, is nearly universal in the quantum theory/computing literature. Alternatively, these coefficients can be represented using *sparse distributed representations* (SDR), wherein each coefficient is represented by a small *subset* of an overall population of representational units and the subsets can *overlap*. Specifically, I consider an SDR model in which the overall population consists of *Q* clusters, each having *K* binary units, so that each coefficient is represented by a set of *Q* units, one per cluster. Thus, $K^Q$ coefficients can be represented with *KQ* units. We can then consider the particular world state, X, whose coefficient's representation, R(X), is the set of *Q* units active at time *t* to have the maximal probability and the probabilities of all other states, Y, to correspond to the size of the intersection of R(Y) and R(X). Thus, R(X) simultaneously serves both as the representation of the particular state, X, and as a *probability distribution over all states*. Thus, set intersection may be used to classically implement quantum superposition. If algorithms exist for which the time it takes to store (learn) new representations and to find the closest-matching stored representation (probabilistic inference) remains *constant* as additional representations are stored, this would meet the criterion of quantum computing. Such algorithms, based on SDR, have already been described. They achieve this "quantum speed-up" with no new esoteric technology, and in fact, on a single-processor, classical (Von Neumann) computer.

**NeuroQuantology 2012; 2:311-315**

A fundamental concept of quantum theory (QT) is that a physical system exists as a superposition of *all* of its possible states and that the act of observation causes exactly one of those states to manifest physically (i.e., collapses the superposition). The number of possible states is exponential in the number of fundamental entities comprising the system.

The *strength of presence* of each possible state in the superposition—*i.e.*, the probability with which it would be observed if measured—is represented by its *probability amplitude coefficient*. It appears to be universally the case in mathematical descriptions of QT thus far, that these coefficients are represented *separately* from each other, as in the standard *N*-qubit register formula, shown in Eq. 1 for *N=2*, where the $\alpha$'s are these coefficients.

$$|\psi\rangle = \alpha_{00}|00\rangle + \alpha_{01}|01\rangle + \alpha_{10}|10\rangle + \alpha_{11}|11\rangle \qquad (1)$$

Presumably, existing computer simulations of QT respect this separateness by representing the coefficients physically disjointly from each other, *i.e.*, in separate computer memory locations. This is referred to as a *localist* representation. It is crucial to understand that in this case, any *single* atomic machine operation (read or write) can affect *only one* memory location and therefore only

---

Corresponding author: Gerard J. Rinkus
Address: Brandeis University, Biology, 405 South St., Waltham MA,
Phone: 617-795-5977
✉ rod.rinkus@gmail.com

one coefficient. Under a localist representation, updating the exponential number of probability amplitude coefficients that characterize the state of a quantum system requires an exponential number of atomic machine operations.

By analogy with QT, a quantum computer (QC) is one whose current representational state is a superposition of all of its possible representational states. If these states, or *codes*, represent the possible states of some observed/modeled world, then the strength of activation of a code can be viewed as representing the probability that the corresponding world state exists and the set of activation strengths of all codes can be viewed as representing the probability distribution over all world states. It appears that the localist representation of such probabilities in QT has carried over universally throughout the QC literature thus far. Consequently, existing QC models have the limitation that atomic operations affect only single probabilities and therefore, that updating the entire probability distribution over the exponential number of represented states requires an exponential number of atomic operations. The requirement for *simultaneously* updating, or evaluating, an exponential number of states is considered a huge, "almost insurmountable" technological hurdle (Hagar, 2008). *"The only general purpose way believed to exist for a classical computer to simulate a quantum computer is to perform a separate computation for each of the exponentially many computations that the quantum computer is doing in parallel [my italics]"* (Perimeter Institute, 2008), a sentiment expressed widely in the QT literature, beginning with (Feynman, 1982).

However, *sparse distributed representations* (SDR) do not have this limitation. In SDR, each represented entity (e.g., each world state, or its probability), X, is represented by a *subset*, R(X), of low-level, e.g., binary, representational units chosen from a much larger population of units. Because subsets may in general overlap, any given unit will generally be included in multiple subsets, i.e., in the codes of multiple states. This can be seen in the particular SDR framework of Figure 1, which consists of $Q=6$ winner-take-all (WTA) clusters, each with $K=3$ units. The convention used here is that each *code* consists of one unit in each cluster. Thus, an exponential number, $K^Q=729$, of codes are possible. For illustration purposes, I chose code A randomly and codes B-G to have decreasing intersection with A (intersecting units are red). Crucially, if any particular code, *e.g.*, R(A), is fully active then *all other* codes can also be considered to be *partially* active in proportion to their intersection with R(A). The bar graphs show activation strength distributions given four different maximally active codes, R(A), R(B), R(D), and R(G). Thus, the fraction of a code's units that are active can be interpreted as representing the probability that the corresponding state exists.

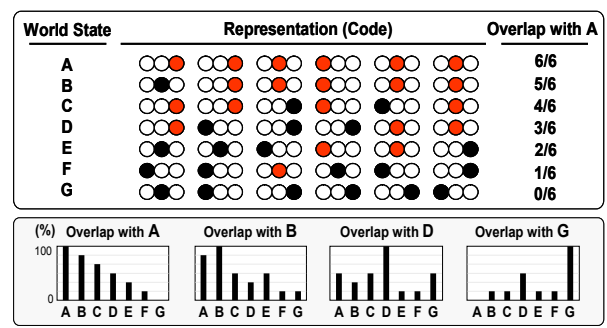

**Figure 1.** A sparse distributed representation (SDR) in which the coding field consists of $Q=6$ clusters, each having $K=3$ binary units, and a code is a set of $Q$ units, one per cluster. When any given code, *e.g.*, R(A), is active, *all* other codes stored in the model are also physically active in proportion to their intersection with R(A). Thus, SDR provides a classical realization of quantum superposition in which probability amplitudes are represented directly and implicitly by sizes of intersections.

Figure 1 suggests the *possibility* of representing superpositions of codes (*i.e.*, representations of world states, *i.e.*, concepts) with SDR. However, to qualify as useful quantum computation, we must show that the computer's dynamics, *i.e.*, the algorithm:

- A. updates the probabilities of all codes (world states) in the superposition in a way that is consistent with the modeled domain's (*i.e.*, world's) dynamics / logic, and

B. updates all those codes simultaneously and immediately, or more precisely, with a number of computing operations that is independent of the number of codes (hypotheses) stored.

For example, a QC that stores a database of images should be able to return the best matching image (or a ranked list of matches) given a query in the same time whether the database stores one image or one billion images. The crucial property necessary to allow an SDR-based approach to meet both constraints A and B is that similar inputs (world states) be mapped to similar (more highly intersecting) codes (SISC). I have developed a classical algorithm that possesses SISC and creates and manipulates sparse distributed codes in accord with these constraints (Rinkus, 1996; 2010). Paraphrasing Pitowsky (2002), for computation to be enhanced by quantum mechanics, models must construct 'clever' superpositions that increase the probability of successfully retrieving the desired, *i.e.,* exact- or closest-matching, result far beyond chance. I believe that the sparse code overlap structure produced by my algorithm, which respects similarity in the input space (*i.e.,* SISC), constitutes such a 'clever superposition'.

The notion of *quantum superposition*, in which all possible states somehow simultaneously *exist* even though only a single state is *physically* observed at any particular instant has long resisted classical interpretation. The Copenhagen School simply asserts that no such interpretation exists, *i.e.,* that quantum superposition has no classical analog, whereas Everett reconciled the quantum and the classical by constructing a reality composed of an astronomical number of classical parallel universes (Everett, 1957), which is an equally unsatisfactory explanation (Garrett, 1993). I believe that the fundamental barrier to achieving a clear classical explanation of superposition, in both QT and QC, has been the virtually ubiquitous use/assumption of localist representations and that achieving such a classical understanding not just of superposition but of all quantum phenomena, *e.g.,* entanglement, requires a move to *distributed representations* (for QC) and to entities that fundamentally have non-zero extension, *i.e.,* non-point masses (for QT).

One example is the approach based on geometric algebra (GA) being developed by Aerts and colleagues (Aerts and Czachor, 2008; Aerts *et al.,* 2009; Patyk-Lonska *et al.,* 2011), which has been formally related to the fundamentally *distributed*, though not necessarily sparse distributed, *reduced representation* models, *e.g.,* (Kanerva, 1994; Plate, 1994; Rachkovskij and Kussul, 2001). In the GA-based theory, codes are formalized as *geometric objects*, which are fundamentally extended, rather than as vectors, which are fundamentally representable as points (with zero extension). By analogy, notice that the set intersections (subsets) central to the SDR-based approach described herein also fundamentally have extent (and cannot be negative: *i.e.,* there is no concept of a "negative subset"). The connection between their GA-based and my SDR-based (essentially, set-based) approach seems a strong candidate for further exploration. However, the main point of this short paper is to explain how it is that sparse distributed representations, in particular, can deliver an exponential speed-up over localist representations.

With SDR, a *separate computation is no longer needed for each represented state*. A separate computation *is* needed for each representational unit, but crucially, and as noted above, this number remains fixed as additional represented states (concepts, memories) are added to the memory. Since any unit participates in the codes of many states, executing a *single* operation on any given unit will simultaneously affect (*i.e.,* perform computational work upon) all of the states in whose codes it participates. Collectively, upon iterating over all units, *all* stored world states will be fully updated. The key point is that in an SDR framework, the number of units remains fixed as the number of represented world states (or essentially, their probabilities) grows. If that framework also possesses SISC then the number of computational steps needed to find the closest-matching stored world state, and more generally achieving property A above, remains fixed as the number of stored world states increases.

The use of SDR allows a *crucial shift* from representing probability amplitudes (i.e., world state probabilities) *explicitly as independently represented scalar coefficients* as in Equation 1, to representing them *implicitly by (sizes of) intersections*. As apparent in Figure 1, even though the underlying units in our model are binary, the number of different probability levels (ranks) represented is $Q+1$, since $Q+1$ intersection sizes are possible between any two codes. Thus, if the model contained $Q=256$ clusters, it would function as if the probabilities represented by the stored codes were 257-valued. The reader might ask: if these probabilities are represented only implicitly in the pattern of intersections between codes, how can they be communicated forward in time to the next step of a computation? They are communicated via a recurrent associative weight matrix, H, which connects the units comprising the coding field back onto themselves, as in Figure 2. The information as to the probabilities of all stored codes at time $t$ is implicit in the pattern of input summations (via the H matrix) for the units, at $t+1$. Yes, it is true that following a decision process, described in (Rinkus, 1996; 2010), one particular code will become *fully* active at $t+1$. However, just as the single code that is fully active at $t$ simultaneously functions as both:

1. the code of one particular state, *i.e.,* the "collapsed superposition", and
2. the probability distribution over all stored codes (states), *i.e.,* the full superposition,

so too does the single code that becomes active at $t+1$, or any other $t$, function in both ways. Thus, the H matrix governs the evolution through time of the superposition. The algorithm by which H is incrementally learned over time by observing the input domain, so that it comes to reflect its dynamics (point B, above) is also given in (Rinkus, 1996; 2010).

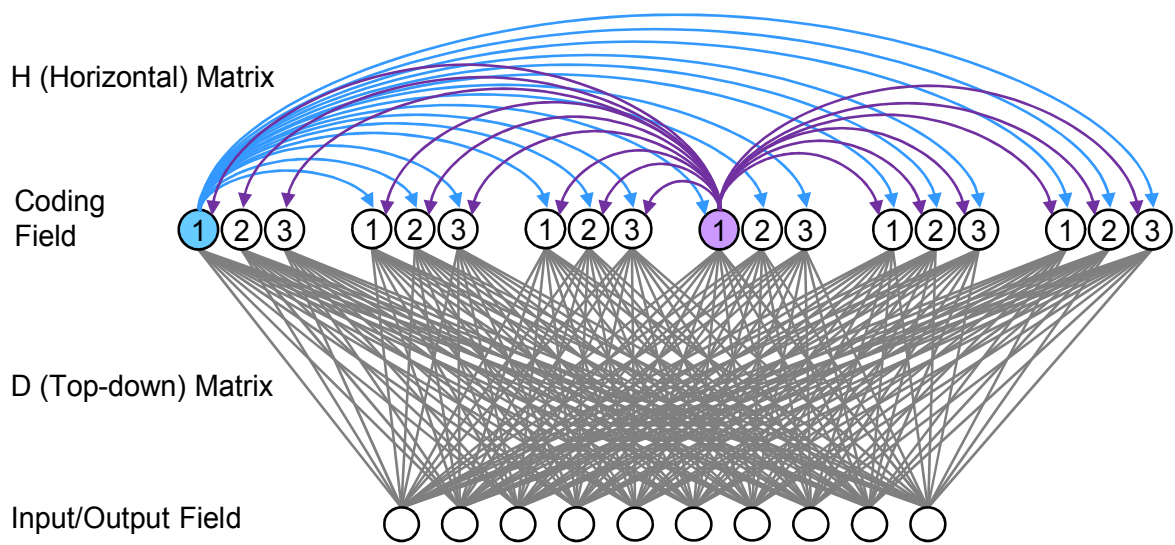

**Figure 2.** The H matrix mediates signals from the single fully active code at $t$, which arrive back at the coding field at $t+1$ and influence which single code becomes active at $t+1$. But, since any single SDR code can also be viewed as a probability distribution over all codes, the H matrix simultaneously mediates the transformation (evolution) from one probability distribution to the next. We show only a portion of the H matrix, namely the output connections from the two highlighted units. Any number of other weight matrices may be connected to/from the coding field, in particular, the D matrix shown here, which allows readout of the coding field at each $t$.

Any number of other weight matrices can also be connected to/from the coding field. In particular, the D matrix shown here, would allow readout of the coding field at each *t*. We can imagine another decision process that occurs over the units of the input/output field at *t*, which causes *one* particular representation to become active in that field. There is no need for that representation to also simultaneously function as a probability distribution over all codes because the coding field already maintains that information. Thus, there is no need for the input/output field to use SDR. Overall, the architecture and suggested dynamics (again, the specific algorithms are described elsewhere) of Figure 2 allows for both the collapse of the superposition at each *t* (the observation being the code that becomes active in the input/output field) and continuous evolution of the superposition.

The question arises: is there anything to gain in moving from binary units to real-valued units, *e.g.,* as 8-byte floats? Clearly, this would vastly increase the number of different probability levels representable by codes consisting of *Q* units. However, I would suggest that the hypothesis spaces relevant to most naturalistic human cognition/decision processes require/entail only a small number of (relative or absolute) probability levels, so that the Q-based implementation of probability levels described herein would suffice. Nevertheless, the essential benefit of representing probabilities implicitly would remain in this case, as would the explanation of how a classical machine implements both superposition collapse at each *t* and continuous evolution of the superposition. Thus, this is a potential future research topic.

I believe that SDR constitutes a classical instantiation of quantum superposition and that switching from localist representations to SDR, which entails no new, esoteric technology, is *the* key to achieving quantum computation in a *single-processor*, classical (Von Neumann) computer.